\documentclass[11pt,a4paper]{article}
\usepackage[hyperref]{acl2017}
\usepackage{times}
\usepackage{latexsym}

\usepackage{graphicx}
\usepackage{color}
\usepackage{multirow}
\usepackage{soul}
\usepackage{todonotes}
\usepackage{url}
\usepackage{booktabs}
\usepackage{tabularx}

\aclfinalcopy 

\title{Creation of an Annotated Corpus of Spanish Radiology Reports}

\author{
        {\bf Viviana Cotik$^{\Box}$}, {\bf Dar\'io Filippo$^{\triangle}$} , {\bf Roland Roller$^{\star}$}, {\bf Hans Uszkoreit$^{\star}$} and {\bf Feiyu Xu$^{\star}$}\\
	    $^{\Box}$Departamento de Computaci\'on, FCEyN, UBA, Argentina\\
        {\tt vcotik@dc.uba.ar} \\
        $^{\triangle }$Hospital De Pediatr\'ia, Prof. Dr. Juan P. Garrahan, Argentina\\
        {\tt dfilippo@garrahan.gov.ar} \\
        $^{\star}$Language Technology Lab, DFKI, Berlin, Germany\\
	    {\tt \{firstname.surname\}@dfki.de} \\}
        
\date{}

\begin{document}
\maketitle
\begin{abstract}

This paper presents a new annotated corpus of 513 anonymized radiology reports written in Spanish. Reports were manually annotated with entities, negation and uncertainty terms and relations. 
The corpus was conceived as an evaluation resource for named entity recognition and relation extraction algorithms, and as input for the use of supervised methods. Biomedical annotated resources are scarce due to confidentiality issues 
and associated costs. 
This work provides some guidelines that could help other researchers to undertake similar tasks. 

\end{abstract}

\section{Introduction}\label{sec:introduction}

The availability of annotated corpora from the biomedical domain, in particular for non-English texts, is scarce. 
There are two main reasons for that: 
the generation of new annotated data is expensive 
due to the need of expert knowledge and 
to privacy issues: the patient and the physician should not be identified from the texts.  
So, although the availability of annotated data is a highly valuable asset for the research community, it is very difficult to access it.

We are interested in supporting physicians with automatic text processing methods, such as named entity recognition (NER), relation extraction (RE), and negation and uncertainty detection in Spanish radiology reports. The extraction of entities and relations from the reports could suggest possible medical problems, that might lead to surgical interventions, such as seen in 
\newcite{Do:2013}, \newcite{Morioka:2016} and \newcite{Lakhani:2009}. 

To the best of our knowledge, there are no publicly available annotated datasets of Spanish medical reports for these tasks.  
For this reason, this work focuses on creating an annotated corpus of Spanish radiology reports. There are some datasets available for other languages in the clinical domain, eg. for English \cite{Uzuner:2011:JAMIA,Pradhan:2013:CLEF,Pradhan:2014:amia}, Swedish \cite{Skeppstedt:2014}, French \cite{Neveol:2015:CLEF}, Polish \cite{Mykowiecka:2009} and German \cite{Roller:2016}. \newcite{Oronoz:2015:corpus}  presented an annotated dataset in Spanish for adverse drug reactions analysis. 



There are different kind of medical reports. 
In our case, reports are very short, %
sentences are not always well formed and 
many of them have a telegraphic style. They contain spelling mistakes and the use of non-standard abbreviations and acronyms 
is frequent. This, added to the use of specialized language of the medical domain, makes the annotation task difficult. 



This work describes the annotation schema, the main guidelines and a brief analysis of the resulting corpus. We are evaluating the possibility of releasing the dataset publicly.



\section{Annotation process}\label{sec:methods}



We developed an annotation guideline, which we improved with three iterations of a process consisting of annotation and revision of doubts in the criteria.
A set of 513 different kinds of ultrasound reports (e.g. kidney, abdominal, small parts)\footnote{Ultrasound reports are texts describing what has been observed in a type of imaging study called ultrasound examination.} 
that were written in a hospital in Argentina were selected for annotation. 
They contain only one section that includes findings, conclusions and suggestions. The reports were anonymized by removing the date of the study, the report number and the patient identification number. Additionally, information about the physicians that performed the study was removed. Regular expressions have been used for this purpose considering the different ways of writing the title of the physicians (e.g. \emph{DR}, \emph{Dr.}, \emph{doctor}, \emph{Dra.}), the doctor's names, the enrollment numbers and the order among these terms. Also names of the doctors appearing with titles or enrollments were searched to see if they appeared without titles and without enrollments and were removed.

The main entities and characteristics that were annotated are presented in Table \ref{table:entities}. An example is included for each case. Abbreviations and acronyms were only annotated for \textit{anatomical entities} and \textit{findings}. 
Table \ref{table:relations} shows the main relations annotated. 
This table exhibits the name of the relation and the entities involved in it.  The relation \textit{occurs in}, for instance, is always constructed between a \textit{finding} and \textit{anatomical entity} and explicits in which anatomical entity the finding occurred. A \emph{texture} can be related by the \emph{texture} relation to a \textit{finding} or to an \textit{anatomical entity}. 

\begin{table}[!th]
\center
\begin{tabular}{ll}
\toprule
\textbf{Name} & \textbf{Example}  \\ 
\midrule
findings (FI) & \textit{cyst}  \\ 
anatomical entities (AE) & \textit{liver}  \\ 
location in body (LO) & \textit{apical}  \\ 
measure (ME) & \textit{0.3 mm}  \\ 
type of measure (TM) & \textit{longitudinal}  \\ 
texture (TE) & \textit{homogeneous}  \\ 
negation terms (NT)  & \textit{has not been detected}  \\ 
uncertainty terms (UT) & \textit{might indicate}  \\ 
abbrev. and acronyms& \textit{RK for right kidney} \\ 
temporal terms (TT) & \textit{preoperative}  \\ 
conditional terms (CT) & \textit{if he has fever again} \\ 
\bottomrule
\end{tabular}
\caption{Overview of entities and characteristics to be annotated.}
\label{table:entities}
\end{table}

\begin{table}[!th]
\center
\begin{tabular}{lll}
\toprule
\textbf{Name} & \textbf{Entity 1} & \textbf{Entity 2} \\ 
\midrule
occurs in & FI & AE  \\ 
located in & LO & AE  \\ 
measure of & ME or TM & AE, LO or FI   \\ 
has measure type & TM & ME \\ 
texture & TE & FI or AE \\ 
negates & NT & FI   \\ 
speculates & UT & FI \\ 
not present & TT or CT & FI \\ 
\bottomrule
\end{tabular}
\caption{Overview of relations to be annotated and the entity types related by them.}
\label{table:relations}
\end{table}

Reports were annotated according to the following main guidelines:

\begin{enumerate}  
\item Annotate the largest possible term. For example, the anatomical entity \emph{retroperitoneo vascular} has as substrings \emph{peritoneo} and \emph{retroperitoneo}, which are also AE. Only \emph{retroperitoneo vascular} has to be annotated.  
\item Use RadLex\footnote{\url{http://radlex.org/}} and UMLS\footnote{\url{https://www.nlm.nih.gov/research/umls/}} to solve doubts. In various cases information might be ambiguous and difficult to narrow down to a particular label. In this case the definition and classification (according to semantic types) in the given ontologies can help. 
\item Annotate concepts with spelling errors. If the meaning of the word is recognizable, the annotation should be carried out as if the word would have been written correctly.
\item Try to keep annotations simple.
\item Annotate relations across sentences.
\item Annotate abbreviations and acronyms  corresponding to AE or FI as abbreviations (or acronyms) and as entities.
\item Prioritize AE over LO in case of doubts.
\item Annotate NT and UT only if there is a relation among them and a FI.
\item Annotate AEs although there is no relation among them and a FI.\footnote{In "right lobe of the liver has the usual size", "right lobe of the liver" should be annotated, although it is not associated to any finding.}
\end{enumerate}

\begin{figure}[ht]
  \includegraphics[width=0.49\textwidth]{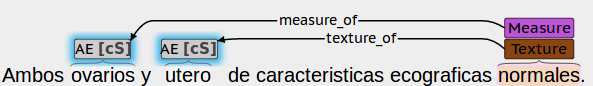}
 \caption{Annotated Text Example-1. \emph{Both ovarys and uterus of normal echographic  signs.}}\label{fig:example}
\end{figure}

\begin{figure}[ht]
  \includegraphics[width=0.49\textwidth]{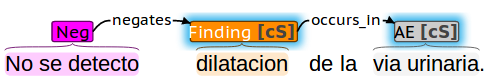}
 \caption{Annotated Text Example-2. \emph{No dilatation of the urinary tract has been detected.}}\label{fig:example2}
\end{figure}

Finally, a concept, that we called \emph{multisegment term}, was introduced: constructions like \emph{intra and extrahepatic} had to lead to the annotation of the entities \emph{intrahepatic} and \emph{extrahepatic}.  
Examples of our annotated corpus are given in Figure \ref{fig:example} and \ref{fig:example2}.

In order to decrease the annotation time, entities, negation and uncertainty terms were pre-annotated automatically. Therefore, regular expressions, UMLS and a manually-created dictionary were used. 
Based on the annotation guideline, two native speakers of Spanish annotated the pre-annotated reports using brat
\footnote{\url{http://brat.nlplab.org/}} \cite{Stenetorp:2012}. Annotations wrongly made by the pre-annotation tool were corrected and missing concepts were included. Relations were introduced by the annotators. Overall, annotators worked for approximately a total of 160 hours on the generation of the corpus.\footnote{All the annotations\-revision iterations were taken into account.} In addition to that, the authors had various discussions to define the final annotation schema based on previous annotations revisions and annotators doubts. We assume that with a stable  annotation schema and once the annotators have less doubts, the annotation process would be quicker.  

\section{Dataset Analysis}
\label{sec:datasetanalysis}

Among other entities, 4398 (405 different\footnote{From now on we omit the word different.}) AE, 2637 (745) FI, 1489 (51) NT and 109 (26) UT have been annotated.  There are 2161 (750)  \emph{occurs in} and 1478 (164) \emph{negates}, among other relations. There appear 470 abbreviations or acronyms corresponding to AE and 7 to FI. 7.89\% (867 out of a total of 10987) of the relations are across-sentence relations. 
The inter-annotator agreement (IAA) for the final annotation schema is 0.89. It was calculated for the 61 reports annotated by both annotators on a token level using the Cohen's Kappa coefficient ($\kappa$) \cite{Cohen:1960}.

\section{Discussion and Conclusions}\label{sec:discussion}

We presented a manually annotated corpus of entities and relations in radiology reports written in Spanish. 
The goal was twofold: to have an annotated dataset available for evaluating NER and RE algorithms results and for training of supervised models and to present an annotation guideline that can be used by other researchers with similar needs. 
The creation of the corpus was not an easy task. Many annotation-revision iterations had to be performed in order to arrive to a stabilized annotation schema. According to what we expected, $\kappa$ improved in each annotation iteration step.  
Data had to be anonymized. 
Furthermore, the shortness of the texts, the abundance of abbreviations and acronyms (about 6\% of the AE and FI are written as such 
and there are 105 different abbreviations or acronyms in 513 reports), the specificity of the medical language, the existence of multi-segment terms 
and the existence of relations between sentences, makes not only the NER and RE tasks, but also the annotation task a difficult one. 

The relation of findings with temporal terms, negation terms and uncertainty terms should be taken into account to determine their factuality. 
The abundance of negated findings 
(56\%) might lead to the implementation of methods to detect negated findings in reports (see \newcite{negex:2001}, and \newcite{CotikACLNegExSpanish} for Spanish). 
The difference of criteria among the annotators helps us determine that the evaluation of NER systems is not an easy task.

Finally, the pre-annotation helped us speed the annotation process, although it might have biased the annotation results. 

\bibliography{acl2017}

\begin{thebibliography}{}
\expandafter\ifx\csname natexlab\endcsname\relax\def\natexlab#1{#1}\fi

\bibitem[{Chapman et~al.(2001)Chapman, Bridewell, Hanbury, Cooper, and
  Buchanan}]{negex:2001}
Wendy Chapman, W.~Bridewell, P.~Hanbury, GF. Cooper, and BG. Buchanan. 2001.
\newblock A simple algorithm for identifying negated findings and diseases in
  discharge summaries.
\newblock {\em J Biomed Inform.\/} 34(5):301--310.

\bibitem[{Cohen(1960)}]{Cohen:1960}
Jacob Cohen. 1960.
\newblock A coefficient of agreement for nominal scales.
\newblock {\em Educational and Psychological Measurement\/} 20(1):37--46.

\bibitem[{Cotik et~al.(2015)Cotik, Stricker, Vivaldi, and
  Rodriguez}]{CotikACLNegExSpanish}
Viviana Cotik, Vanesa Stricker, Jorge Vivaldi, and Horacio Rodriguez. 2015.
\newblock {Syntactic methods for negation detection in radiology reports in
  Spanish}.
\newblock In {\em ACL - Workshop on Replicability and Reproducibility in
  Natural Language Processing: adaptative methods, resources and software\/}.
  Buenos Aires, Argentina.

\bibitem[{Do et~al.(2013)Do, Wu, Maley, S., and Biswal}]{Do:2013}
B.~Do, A.S. Wu, J.~Maley, S., and Biswal. 2013.
\newblock Automatic retrieval of bone fracture knowledge using natural language
  processing.
\newblock {\em J Digit Imaging\/} 26(4):709--13.

\bibitem[{Lakhani and Langlotz(2009)}]{Lakhani:2009}
Paras Lakhani and Curtis~P. Langlotz. 2009.
\newblock Automated detection of radiology reports that document non-routine
  communication of critical or significant results.
\newblock {\em J Digit Imaging\/} 23(6):647--57.

\bibitem[{Morioka et~al.(2016)Morioka, Meng, Taira, Sayre, Zimmerman,
  Ishimitsu, Huang, Shen, and El-Saden}]{Morioka:2016}
C.~Morioka, F.~Meng, R.~Taira, J.~Sayre, P.~Zimmerman, D.~Ishimitsu, J.~Huang,
  L.~Shen, and S.~El-Saden. 2016.
\newblock {Automatic Classification of Ultrasound Screening Examinations of the
  Abdominal Aorta.}, journal = {J Digit Imaging}, volume = {29}, number = {6},
  pages = {742--48} .

\bibitem[{Mykowiecka et~al.(2009)Mykowiecka, Marciniak, and
  Kup\'{s}\'{c}}]{Mykowiecka:2009}
Agnieszka Mykowiecka, Ma\l{}gorzata Marciniak, and Anna Kup\'{s}\'{c}. 2009.
\newblock Rule-based information extraction from patients' clinical data.
\newblock {\em Journal of Biomedical Informatics\/} 42(5):923 -- 936.
\newblock Biomedical Natural Language Processing.

\bibitem[{N{\'{e}}v{\'{e}}ol et~al.(2015)N{\'{e}}v{\'{e}}ol, Grouin, Tannier,
  Hamon, Kelly, Goeuriot, and Zweigenbaum}]{Neveol:2015:CLEF}
Aur{\'{e}}lie N{\'{e}}v{\'{e}}ol, Cyril Grouin, Xavier Tannier, Thierry Hamon,
  Liadh Kelly, Lorraine Goeuriot, and Pierre Zweigenbaum. 2015.
\newblock {CLEF} ehealth evaluation lab 2015 task 1b: Clinical named entity
  recognition.
\newblock In {\em Working Notes of {CLEF} 2015 - Conference and Labs of the
  Evaluation forum, Toulouse, France, September 8-11, 2015.\/}.

\bibitem[{Oronoz et~al.(2015)Oronoz, Gojenola, P{\'e}rez, de~Ilarraza, and
  Casillas}]{Oronoz:2015:corpus}
Maite Oronoz, Koldo Gojenola, Alicia P{\'e}rez, Arantza~D{\'i}az de~Ilarraza,
  and Arantza Casillas. 2015.
\newblock On the creation of a clinical gold standard corpus in spanish: Mining
  adverse drug reactions.
\newblock {\em Journal of Biomedical Informatics\/} 56:318 -- 332.

\bibitem[{Pradhan et~al.(2014)Pradhan, Elhadad, South, Martinez, Christensen,
  Vogel, Suominen, Chapman, and Savova}]{Pradhan:2014:amia}
Sameer Pradhan, No{\'e}mie Elhadad, Brett~R South, David Martinez, Lee
  Christensen, Amy Vogel, Hanna Suominen, Wendy~W Chapman, and Guergana Savova.
  2014.
\newblock Evaluating the state of the art in disorder recognition and
  normalization of the clinical narrative.
\newblock {\em Journal of the American Medical Informatics Association\/} .

\bibitem[{Pradhan et~al.(2013)Pradhan, Elhadad, South, Mart{\'{\i}}nez,
  Christensen, Vogel, Suominen, Chapman, and Savova}]{Pradhan:2013:CLEF}
Sameer Pradhan, No{\'{e}}mie Elhadad, Brett~R. South, David Mart{\'{\i}}nez,
  Lee~M. Christensen, Amy Vogel, Hanna Suominen, Wendy~W. Chapman, and
  Guergana~K. Savova. 2013.
\newblock {Task 1: ShARe/CLEF eHealth Evaluation Lab 2013}.
\newblock In {\em Working Notes for {CLEF} 2013 Conference , Valencia, Spain,
  September 23-26, 2013\/}.

\bibitem[{Roller et~al.(2016)Roller, Hans~Uszkoreit, Seiffe, Mikhailov, Staeck,
  Budde, Halleck, and Schmidt}]{Roller:2016}
Roland Roller, Feiyu~Xu Hans~Uszkoreit, Laura Seiffe, Michael Mikhailov, Oliver
  Staeck, Klemens Budde, Fabian Halleck, and Danilo Schmidt. 2016.
\newblock {A fine-grained corpus annotation schema of German nephrology
  records}.
\newblock {\em Proceedings of the Clinical Natural Language Processing
  Workshop\/} 28(1):69--77.

\bibitem[{Skeppstedt et~al.(2014)Skeppstedt, Kvist, Nilsson, and
  Dalianis}]{Skeppstedt:2014}
Maria Skeppstedt, Maria Kvist, Gunnar~H. Nilsson, and Hercules Dalianis. 2014.
\newblock Automatic recognition of disorders, findings, pharmaceuticals and
  body structures from clinical text: An annotation and machine learning study.
\newblock {\em Journal of Biomedical Informatics\/} 49:148 -- 158.

\bibitem[{Stenetorp et~al.(2012)Stenetorp, Pyysalo, Topi\'{c}, Ohta, Ananiadou,
  and Tsujii}]{Stenetorp:2012}
Pontus Stenetorp, Sampo Pyysalo, Goran Topi\'{c}, Tomoko Ohta, Sophia
  Ananiadou, and Jun'ichi Tsujii. 2012.
\newblock {brat}: a web-based tool for {NLP}-assisted text annotation.
\newblock In {\em Proceedings of the Demonstrations Session at {EACL} 2012\/}.
  Association for Computational Linguistics, Avignon, France.

\bibitem[{Uzuner et~al.(2011)Uzuner, South, Shen, and
  DuVall}]{Uzuner:2011:JAMIA}
{\"O}zlem Uzuner, Brett~R South, Shuying Shen, and Scott~L DuVall. 2011.
\newblock {2010 i2b2/VA challenge on concepts, assertions, and relations in
  clinical text}.
\newblock {\em Journal of the American Medical Informatics Association\/}
  18(5):552--556.

\end{thebibliography}
\bibliographystyle{acl_natbib}

\appendix

\end{document}